\documentclass{article}

\usepackage{graphicx}  

\usepackage[preprint]{neurips_2025}

\usepackage[utf8]{inputenc} 
\usepackage[T1]{fontenc}    
\usepackage{hyperref}       
\usepackage{url}            
\usepackage{booktabs}       
\usepackage{amsfonts}       
\usepackage{nicefrac}       
\usepackage{microtype}      
\usepackage{xcolor}         
\usepackage{wrapfig}
\usepackage{tabularx}  
\usepackage{booktabs}
\usepackage{multirow}
\usepackage{tabularx}
\usepackage{array}
\usepackage{caption}
\usepackage{bm}
\usepackage{algorithm}
\usepackage{amsmath}
\usepackage{algorithmic}
\usepackage{subfigure}
\title{HyP-ASO: A Hybrid Policy-based Adaptive Search Optimization Framework for Large-Scale Integer Linear Programs}

\author{%
  Ning Xu \\
  School of Mathematics and Statistics \\
  Nanjing University of Science and Technology \\
  Nanjing, China \\
  \texttt{nxu@njust.edu.cn} \\
  \And
  Junkai Zhang \\
  Institute of Automation \\
  Chinese Academy of Sciences \\
  Beijing, China \\
  \And
  Huigen Ye \\
  Department of Computer Science and Technology \\
  Tsinghua University \\
  Beijing 100084, China \\
  \And
  Yang Wu \\
  Institute of Automation \\
  Chinese Academy of Sciences \\
  Beijing, China \\
  \And
  Hua Xu \\
  Department of Computer Science and Technology \\
  Tsinghua University \\
  Beijing 100084, China \\
  \And
  Huiling Xu \\
  School of Mathematics and Statistics \\
  Nanjing University of Science and Technology \\
  Nanjing, China \\
  \And
  Yifan Zhang \\
  Institute of Automation, Chinese Academy of Sciences \\
  Beijing, China \\
  University of Chinese Academy of Sciences \\
  Nanjing, China \\
}

\begin{document}

\maketitle

\begin{abstract}
    Directly solving large-scale Integer Linear Programs (ILPs) using traditional solvers is slow due to their NP-hard nature. While recent frameworks based on Large Neighborhood Search (LNS) can accelerate the solving process, their performance is often constrained by the difficulty in generating sufficiently effective neighborhoods. To address this challenge, we propose HyP-ASO, a hybrid policy-based adaptive search optimization framework that combines a customized formula with deep Reinforcement Learning (RL). The formula leverages feasible solutions to calculate the selection probabilities for each variable in the neighborhood generation process, and the RL policy network predicts the neighborhood size. Extensive experiments demonstrate that HyP-ASO significantly outperforms existing LNS-based approaches for large-scale ILPs. Additional experiments show it is lightweight and highly scalable, making it well-suited for solving large-scale ILPs.
\end{abstract}

\section{Introduction}
\label{Introduction}

Combinatorial Optimization Problems (COPs)  are a class of optimization problems with discrete decision variables and a finite search space \cite{papa1998handbook}. These problems have broad applications in real-world scenarios, including bin packing \cite{Monaci06}, production planning \cite{Yu13}, electronic design automation \cite{Chen24}, and routing \cite{Nazari18, Wu24}. Integer Linear Programs (ILPs) are commonly used to model many COPs \cite{Korte11, Lancia18}. However, solving ILPs is extremely challenging due to their NP-hard nature \cite{Karp72}. Furthermore, as the scale of ILPs increases, the solving cost increases dramatically, yet such large-scale ILPs frequently arise in the real world \cite{Martin98}. Therefore, developing efficient methods for solving large-scale ILPs has become a critical research focus.

Exact algorithms based on Branch and Bound (B\&B) are the most widely used methods for solving ILPs \cite{Korte11}. B\&B systematically explores the solution space by recursively solving branch nodes, each of which corresponds to a sub-ILP \cite{Lawler1966BranchAB}. It is the foundation for commercial solvers like Gurobi \cite{gurobi}, CPLEX \cite{CPLEX}, and the academic solver SCIP \cite{Bolusani2024}. Recent research has explored integrating Machine Learning (ML) with B\&B to exploit distributional similarities across problem instances \cite{Bengio21, Scavuzzo24}. They focus on optimizing key components of solvers, such as primal heuristics \cite{Chmiela2021LearningTS, Nair20}, branch and node selection \cite{Gasse2019ExactCO, He2014LearningTS, Wang2023LearningTB}, and cutting plane selection \cite{Puigdemont2024LearningTR, Wang2023LearningCS}. Despite some advances, efficiently obtaining high-quality feasible solutions for large-scale ILPs remains a significant challenge. The exponential growth of the B\&B tree primarily limits the scalability and efficiency. Moreover, the high training cost of learning-based methods hinders their practical use on large-scale ILPs.

Therefore, some researchers have turned to combining metaheuristics with B\&B from an external perspective rather than directly improving B\&B itself \cite{pmlr-v202-huang23g, Song20, Sonnerat2021LearningAL, Wu2021LearningLN, Yuan25}. These approaches employ the Large Neighborhood Search (LNS) algorithm, a metaheuristic that iteratively improves an initial solution by selecting a subset of variables as the neighborhood to re-optimize while keeping the rest fixed \cite{Pisinger2019LargeNS}. Within this paradigm, two main strategies have emerged. The first imitates expert-designed policies to guide the neighborhood selection process \cite{pmlr-v202-huang23g, Sonnerat2021LearningAL}. It heavily depends on expert knowledge and suffers from limited scalability due to the high cost of collecting expert data for large-scale ILPs. The second leverages Reinforcement Learning (RL) to generate neighborhoods directly \cite{Wu2021LearningLN, Yuan25}. However, it faces significant scalability issues when dealing with large-scale ILPs with high-dimensional discrete action spaces. 

Recently, Ye et al. \cite{ye2023adaptive} introduced the Adaptive Constraint Partition (ACP) method, which enhances neighborhood selection in LNS by partitioning constraints to identify more promising variable subsets. Notably, this represents the first application of an LNS-based approach to large-scale ILPs involving millions of variables and constraints. Building upon ACP, they further proposed two fast optimization frameworks that leverage graph partition, variable reduction, and small datasets for solving large-scale ILPs \cite{Ye24, Ye23}. However, their effectiveness is limited by fixed neighborhood sizes and inflexible variable selection rules.


\begin{figure*}[ht]
    \centering
    \includegraphics[width=1\linewidth]{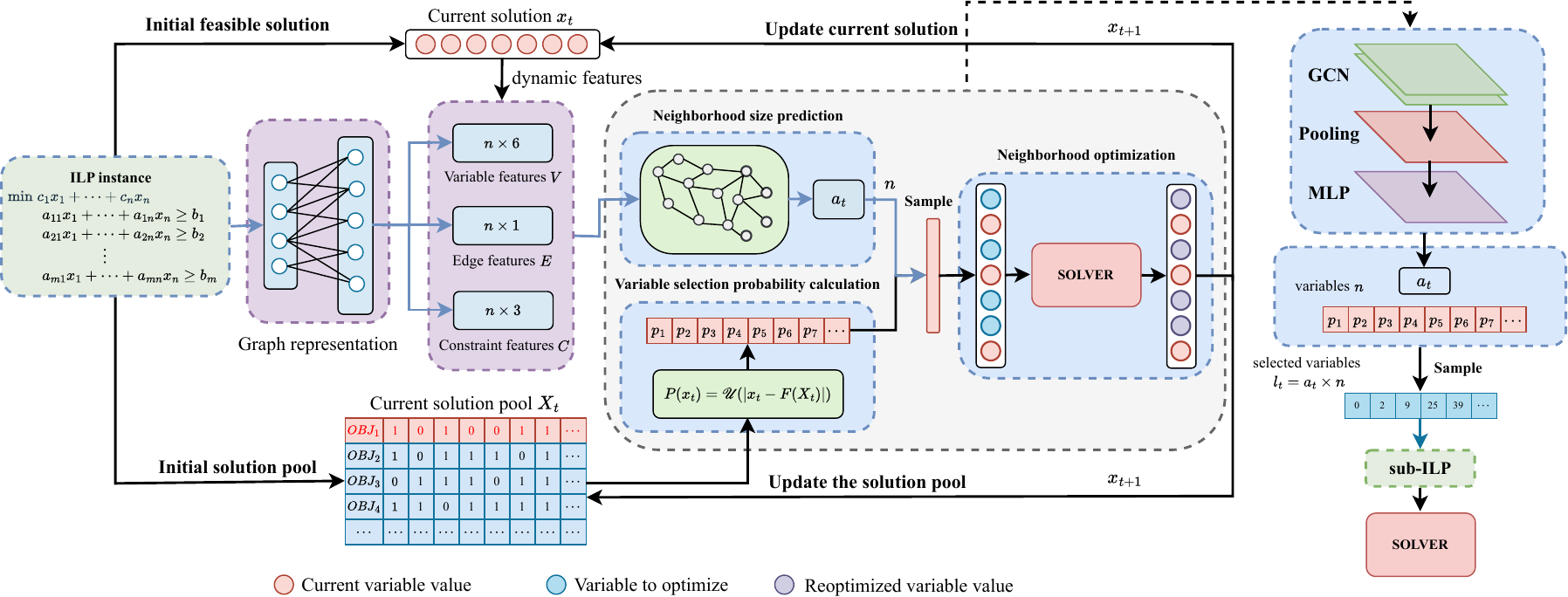}
    \caption{Overview of the HyP-ASO framework: The blue lines illustrate the inference process of the RL policy. For each ILP instance, we initialize both the feasible solution and the solution pool. In each iteration, the neighborhood generation process is divided into three steps. First, a customized formula is used to calculate the selection probability for each variable. Then, an RL policy network predicts the neighborhood size. Based on the calculated probabilities and the predicted size, a sampling method is used to select variables to generate a neighborhood, resulting in a sub-ILP. Finally, this sub-ILP is solved by a solver, completing the neighborhood optimization process. Key details are provided on the right.}
    \label{fig:framework}

\end{figure*}

Motivated by the analysis above, we propose HyP-ASO, a hybrid policy-based search optimization framework that aims to find higher-quality solutions for large-scale ILPs in less time. Our approach combines a customized formula with deep RL to adaptively generate effective neighborhoods for accelerating the solving of large-scale ILPs. Specifically, we first quickly solve the ILP to obtain an initial feasible solution and a solution pool. The pool contains all feasible solutions and their corresponding objective values in the solving process. Then, we implement a dynamic selection mechanism for generating neighborhoods, which involves three key steps: (i) calculating variable selection probabilities based on the solution pool and the customized formula, (ii) employing deep RL to train a policy network that adaptively predicts the neighborhood size, and (iii) using a sampling method to select variables to generate the neighborhood, resulting in a sub-ILP. Finally, we use a solver to solve the sub-ILP, updating the current feasible solution and solution pool. 

The contributions are as follows: 
\begin{itemize}
    \item We design a customized formula that leverages the solution pool to improve the quality of variable selection when generating neighborhoods.
    \item Based on RL, we develop an adaptive mechanism that predicts appropriate neighborhood sizes to improve search efficiency.
    \item The proposed approach performs significantly better than existing methods across multiple large-scale ILP benchmarks.
\end{itemize}
\section{Related Work}
\label{Related Work}

This section provides a brief review of the literature relevant to this paper. We first discuss feasible solutions, which are important for the customized formula. We then present existing learning-based LNS methods for solving ILPs and emphasize the main challenges they encounter.

\textbf{Learning to generate high-quality feasible solutions}

High-quality feasible solutions play a critical role in ILP solving by providing high-quality primal bounds that accelerate the overall solving process. To obtain such solutions efficiently, solvers typically rely on heuristic methods without guaranteeing optimality \cite{Berthold2006}. Recent studies have explored learning-based heuristics to improve this process, aiming to produce higher-quality feasible solutions in less time \cite{Ding2019AcceleratingPS, Han2023AGP, Nair20, Sonnerat2021LearningAL, Yoon2022ConfidenceTN}. These works reveal that high-quality solutions often share structural regularities, with certain variable assignments appearing consistently. Motivated by this, we design a ranking-based variable selection formula that leverages a dynamically updated solution pool and a scoring mechanism \cite{sun2021using} to guide neighborhood generation more effectively.

\textbf{Learning-based LNS methods for solving ILPs}  

LNS algorithms have been proven effective in solving various COPs \cite{ Dumez2021, Feijen2024, Nair2020, Ngoo2024}. Integrating ML with LNS for solving ILPs focuses on improving the quality of selected variables, leading to generating more effective neighborhoods. They can be broadly categorized into three main strategies: (i) training a policy network by RL to directly output a subset of variables as neighborhoods \cite{Nair2020, Wu2021LearningLN, Yuan25}, and (ii) training a policy network by imitation learning to assign scores to variables \cite{pmlr-v202-huang23g, Song20, Sonnerat2021LearningAL}. While most existing methods focus on small-scale ILPs, Ye et al. \cite{ye2023adaptive} proposed the ACP method, which extends LNS to tackle large-scale ILPs. This was further developed into two fast optimization frameworks incorporating graph partitioning and variable reduction techniques \cite{Ye24, Ye23}. However, these approaches still depend on fixed neighborhood size and inflexible variable selection mechanisms, which limit their scalability and generalization. Overall, prior work demonstrates the potential of learning-enhanced LNS, but scalable and adaptive neighborhood generation remains a key challenge for large-scale ILPs.

\section{Preliminaries}
\label{Preliminaries}

This section introduces ILPs, bipartite graph representation, and the LNS algorithm.

\subsection{Integer Linear Programs}

ILP is a powerful method for solving COPs \cite{Lancia18}. It formulates the problem as a linear program with decision variables constrained to integer values, making it especially suited for COPs involving discrete decisions \cite{Chen24, Yu13, Monaci06}. In general, it can be formulated as 
\begin{equation}\label{ilps_format}
\begin{aligned}
\min_{x} ~ &  c^Tx, \\
& Ax \geq b, ~ x \geq 0, ~ x \in \mathbb{Z}^n,
\end{aligned}
\end{equation}
where $x=\left[x_{1}, x_{2}, \cdots, x_{n}\right]^T $ is the vector of the decision variable, $c=\left[c_{1}, c_{2}, \cdots, c_{n}\right]^T \in \mathbb{R}^n$ is the vector of objective coefficients, and $A \in \mathbb{R}^{m \times n}$ and $b \in \mathbb{R}^m$ represent the constraint coefficient matrix and the constraint right-hand side value vector, respectively. This work considers large-scale ILPs, where the number of variables $n$ and constraints $m$ both exceed tens of thousands.

\subsection{Bipartite Graph Representation}\label{bgp}

\begin{figure}
    \centering
    \includegraphics[width=0.6\linewidth]{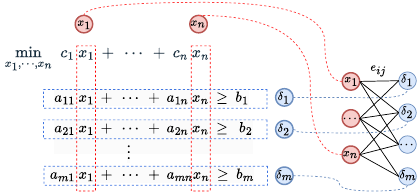}
    \caption{The bipartite graph representation of an ILP instance. The variables $x_{1}, x_{2}, \ldots, x_{n}$ and constraints $\delta_{1}, \delta_{2}, \ldots, \delta_{m}$ represent the left and right nodes of the bipartite graph, respectively. Edges $e_{ij}$ links $i$-th variable and $j$-th constraint.
    }
    \label{fig: Bipartite Graph Representation}
\end{figure}

For each ILP instance, we use a bipartite graph representation $\mathcal{G} = (\mathcal{V}, \mathcal{C}, \mathcal{E})$ to capture its key attributes \cite{Gasse2019ExactCO}. This approach has been widely adopted in works that leverage ML to improve the solving efficiency for ILPs \cite{Puigdemont2024LearningTR, Sonnerat2021LearningAL, Wang2023LearningTB}. Figure \ref{fig: Bipartite Graph Representation} illustrates the relationship between an ILP instance and its corresponding bipartite graph. In the bipartite graph $\mathcal{G}$, we define $\mathcal{V} = \{x_{i}\}$ as the nodes representing decision variables and $\mathcal{C} = \{\delta_{j}\}$ as the nodes representing constraints, where $i = 1, 2, \ldots, n $ and $j=1, 2, \ldots, m$. The edges $\mathcal{E} = \{e_{ij}\}$ connect variables and constraints. Each edge $e_{ij}$ corresponds to the coefficient $a_{ij}$ of the $i$-th variable in the $j$-th constraint. 


The features of $\mathcal{G}$ are denoted by $\mathbf{V}$, $\mathbf{C}$, and $\mathbf{E}$, respectively. They are derived from the coefficients of the ILP instance \cite{Ye23}. To enhance the embedding capacity of the GNN, we introduce random features into the bipartite graph representation \cite{Chen2022OnRM}. This incorporation enriches the representation, improving the learning and generalization capabilities when applying ML to address large-scale ILPs.

\subsection{Large Neighborhood Search}\label{LNS}


LNS is a metaheuristic algorithm \cite{Shaw1998UsingCP} that iteratively improves an initial feasible solution through alternating destroy and repair phases, each forming a solving step. At step $t$, a subset of variables is selected to form a neighborhood $\mathcal{N}(x_t)$ in the destroy phase, resulting in a sub-ILP. This sub-ILP is solved within a time limit to obtain a new feasible solution $x_{t+1}$ in the repair phase. The process continues until a termination condition is met, such as reaching a time limit or a satisfactory objective value, at which point $x_{t+1}$ is returned as the final solution.

The destroy phase plays a key role in the LNS algorithm, as it determines the potential for finding high-quality solutions within the neighborhood search space and the difficulty of solving sub-ILPs. Both factors directly affect the overall solving efficiency. The existing methods incorporating ML into LNS for solving ILPs focus on learning a variable selection strategy to generate better neighborhoods. Our proposed framework is also based on the LNS algorithm and aims to achieve the same goal. Unlike existing methods \cite{pmlr-v202-huang23g, Song20, Sonnerat2021LearningAL, Wu2021LearningLN, Ye24, Ye23, Yuan25}, we decompose the destroy phase into two steps: (i) using a customized formula to calculate the selection probability for each variable based on the solution pool and (ii) employing a policy network to predict the neighborhood size. This method can generate more effective neighborhoods for accelerating the solving process for large-scale ILPs.
\section{Methodology}
\label{method}

In this section, we provide a detailed overview of our proposed approach. We first describe the customized formula to calculate variable selection probability based on solution pools. Then, we explain how to train a policy network to predict the neighborhood size. Finally, we detail the neighborhood optimization process and present the complete algorithm. The overall architecture of the HyP-ASO framework is illustrated in Figure \ref{fig:framework}.

\subsection{Variable Selection Probability Calculation}

For a given ILP instance $\mathcal{Q}$, we quickly solve it to obtain an initial feasible solution $x_{0}$ and an initial solution pool $X_{0}$ within a specified initial solving time $t_{0}$. The pool $X_{0}$ contains all feasible solutions and their corresponding objective values up to the time $t_{0}$.

Assume that there exist $q$ feasible solutions $\mathbf{x}_{t, j}, j = 1, \ldots, q$ in $X_{t}$, we can obtain their quality rankings $r_{j}$ according to their objective values. Let $x_{t} = \left[x_{t, 1}, x_{t, 2}, \cdots, x_{t, n}\right]$ represent the variables at step $t$, and then we calculate the variable selection probability $P(x_{t})$ using a customized formula based on the solution pool $X_{t}$. Before this, we introduce the confidence score of $x_{t, i}$, which means the probability that $x_{t, i}$ equals $1$. This confidence score varies at each step, depending on the solutions in the solution pool. It can be calculated as follows:
\begin{equation}\label{normalization_formula}
f(x_{t,i}) = \frac{\sum_{j=1}^{q}\frac{\mathbf{x}_{t, i, j}}{r_j}}{\max\limits_{i=1,2,\ldots,n}{\sum_{j=1}^{q}\frac{\mathbf{x}_{t, i, j}}{r_j}}},
\end{equation}
where $\mathbf{x}_{t, i, j}$ represents the value of the variable $x_{t, i}$ in the $j$-th feasible solution. The term $\sum_{j=1}^{q}\frac{\mathbf{x}_{t, i, j}}{r_j}$ aggregates the weighted contributions of the variable $x_{t, i}$ across all feasible solutions in the pool $X_{t}$, with higher-quality solutions contributing more due to their smaller rankings $r_{j}$. The denominator in (\ref{normalization_formula}) ensures that confidence scores are scaled relative to the maximum score of all variables, mapping the scores to the interval $[0, 1]$. Then the variable selection probability $P(x_{t}) = \left[p(x_{t,1}), p(x_{t,2}), \cdots, p(x_{t,n})\right]$ are computed as
\begin{equation}\label{customized_formula}
P(x_{t}) = \mathcal{U}(\left|x_{t} - F(X_{t})\right|),
\end{equation}
where $F(X_{t}) = \left[f(x_{t,1}), f(x_{t,2}), \cdots, f(x_{t, n}) \right]$ and $\mathcal{U}(\cdot)$ is the unit sum normalization function, which can ensure that $\sum_{i=1}^{n} p(x_{t, i}) = 1$.

It is noted that if $x_{t, i}=1$, the probability $p(x_{t, i}$) is proportional to $\left|1-f(x_{t, i})\right|$, which emphasizes the deviation of the variable's value from its confidence score. A larger deviation indicates higher uncertainty, suggesting a greater likelihood that the variable needs re-optimization. Conversely, if $x_{t, i}=0$, it is proportional to $\left|0-f(x_{t, i})\right|$. In addition, if $p(x_{t, i})=0$, it is replaced by the smallest positive probability among all variables. Furthermore, if all probabilities are equal, they are uniformly assigned $\frac{1}{n}$, where $n$ is the total number of variables.

The above method prioritizes variables with higher uncertainty for re-optimization and allows for selecting variables with smaller deviations. As a result, it ensures the reliability of the neighborhoods and improves the ability to escape from the local optima.

\subsection{Neighborhood Size Prediction}

After defining the variable selection rule, the next step is determining the neighborhood size, which is also crucial for neighborhood generation using a sampling method. The neighborhood size plays a critical role: if it is too large, the solving time for the sub-ILP may exceed the allowed limit; if it is too small, the neighborhood search space may miss high-quality solutions. Existing methods ignore the dynamic adjustment of the neighborhood size to improve the performance of LNS \cite{pmlr-v202-huang23g, Sonnerat2021LearningAL, Ye24}. In this paper, we employ deep RL to train a policy network to predict the neighborhood size adaptively. As detailed below, this task is framed as a multi-step sequential decision-making process.

\textbf{Reinforcement Learning Formulation}

We model the task as a continuous Markov Decision Process, consisting of the following components: state space $\mathcal{S}$, action space $\mathcal{A}$, transition function, and reward function $\mathcal{R}:\mathcal{S} \times \mathcal{A} \rightarrow \mathbb{R}$. In our framework, the policy for neighborhood size prediction is treated as the agent, while the remaining components of HyP-ASO are the environment. The details are as follows:

\textit{\textbf{State space $\mathcal{S}$.}} The state $s_{t} \in \mathcal{S}$ is a bipartite graph that encodes the information of the ILP instance $\mathcal{Q}$. To capture more dynamic information in the iterative optimization process, the current solution $x_{t}$ is incorporated into $s_{t}$ as a dynamic feature, providing a more comprehensive and adaptive state representation.

\textit{\textbf{Action space $\mathcal{A}$.}} For each state $s_{t}$, the agent outputs a positive value $a_{t}$ within the range $(0, 1)$ as an action. This value $a_{t}$ represents the ratio of the total variable number $n$ in the ILP instance $\mathcal{Q}$. It determines the neighborhood size $l_{t}=a_{t}\times n$ at step $t$.

\textit{\textbf{Transition.}} A sub-ILP is solved by a solver within a specified iteration solving time $t_{p}$ to obtain the updated $x_{t+1}$. Following this, the state $s_{t}$ transitions to $s_{t+1}$ by updating the current solution $x_{t}$ to $x_{t+1}$. The sub-ILP at step $t$ can be formulated as 
\begin{equation}\label{sub-ilps_format}
\begin{aligned}
\min_{x} ~ &  c^Tx, \\
& Ax \geq b, ~ x \geq 0, ~ x \in \mathbb{Z}^n,\\
& x_{i} = x_{t,i}, ~ \forall x_{i}\notin \mathcal{N}(x_t),
\end{aligned}
\end{equation}
where $x = \left[x_{1}, x_{2}, \cdots, x_{n}\right]^T$, $c=\left[c_{1}, c_{2}, \cdots c_{n}\right]^T$ and $\mathcal{N}(x_t)$ denotes the neighborhood, which contains the variables to be re-optimized. $x_{t,i}$ is the value of the variable $x_{i}$ at step $t$, and $x_{i} = x_{t,i}, ~ \forall x_{i}\notin \mathcal{N}(x_t)$ means that we fix the variables $x_{i}$ that are not in the neighborhood $\mathcal{N}(x_t)$. The solution of (\ref{sub-ilps_format}) is denoted by $x_{t+1}$.

\textit{\textbf{Reward function.}} Our method aims to find higher-quality feasible solutions for ILPs in less time. To achieve this, we define the reward function as $r_{t} = \mathcal{R}(s_{t}, a_{t}) = x_{t}-x_{t+1}$. Assuming that the total number of interactions between the agent and the environment within an episode is $T$, the cumulative reward is expressed as

\begin{equation}\label{return}
\begin{aligned}
R = \sum_{t=1}^{T} e^{-k_{r} t^2}r_{t},
\end{aligned}
\end{equation}
where the $k_{r}$ is a designed discount factor. The term $e^{-k_{r}t^{2}}$ emphasizes the significance of earlier stages in the iterative optimization process of LNS, as each iteration builds upon and refines the previous solution.

\textbf{Policy Network}

We learn a policy $\pi_{\theta}(a_{t}|s_{t}): s_{t} \rightarrow a_{t}$ to predict the neighborhood size at each step, where $\theta$ represents the network parameters. The core of the network architecture is a two-layer GNN,  specifically designed for efficient global information aggregation. The details of the network architecture and the training algorithm are as follows:

\textit{\textbf{Network Architecture.}} We use bipartite graphs to represent ILP instances. Then, we employ a fully connected Multi-Layer Perceptron (MLP) layer to map each variable, constraint, and edge node feature in the bipartite graph to the embedding space $\mathbb{R}^{d}$, denoted as $\mathbf{v}_{i}$, $\mathbf{c}_{j}$, $\mathbf{e}_{ij}$, respectively. Then, following the previous work \cite{Gasse2019ExactCO}, we use a GNN with two interleaved half-convolution layers to perform two rounds of message passing for effective global information aggregation. In each round, information is propagated from the variable nodes $\mathbf{v}_{i}$ to the constraint nodes $\mathbf{c}_{j}$, followed by a reverse propagation from the constraint nodes $\mathbf{c}_{j}$ back to the variable nodes $\mathbf{v}_{i}$. The propagation mechanism is expressed as follows:
\begin{equation}
\begin{aligned}
\mathbf{c}_{j}^{k+1} & = f_{c}^{k}\big(\mathbf{c}_{j}^{k}, \sum\limits_{(x_{i}, \delta_{j})\in \mathcal{E}} g_{c}^{k}(\mathbf{c}_{j}^{k}, \mathbf{v}_{i}, \mathbf{e}_{ij})\big), \\
\mathbf{v}_{i}^{k+1} & = f_{v}^{k}\big(\mathbf{v}_{i}^{k}, \sum\limits_{(x_{i}, \delta_{j})\in \mathcal{E}} g_{v}^{k}(\mathbf{v}_{i}^{k}, \mathbf{c}_{j}, \mathbf{e}_{ij})\big),
\end{aligned}
\end{equation}
where $f_{c}^{k}$, $g_{c}^{k}$ $f_{v}^{k}$ and $g_{v}^{k}$ represent MLP layers in the $k$-th GNN layer. They update node embeddings by aggregating messages from neighboring nodes and incorporating edge features $\mathbf{e}_{ij}$. The summation over $\mathcal{E}$ reflects message aggregation across all edges connected to the current node.

After the message-passing phase, variable embeddings are processed through a global pooling layer and then followed by two MLP layers to compute the mean and standard deviation that parameterize the policy distribution $\pi_{\theta}(\cdot|s_{t})$. The action $a_{t}$ is sampled from this distribution and clamped to the range $(0, 1)$, ensuring valid outputs for subsequent decision-making. Details of the network structure are provided in Appendix \ref{network structure}.

\textit{\textbf{Training algorithm.}} In this paper, we utilize Proximal Policy Optimization (PPO) to train the policy network \cite{Schulman2017ProximalPO}. PPO is widely recognized for its high data efficiency and stable performance, particularly in high-dimensional and complex environments. It employs an actor-critic framework, where the actor, represented by the policy network $\pi_{\theta}$, is responsible for action selection, and the critic, represented by the value network $v_{\phi}$, estimates the expected reward for a given state. Both the critic and the actor share the same encoder architecture, with the critic using MLP layers as its decoder to output a scalar value that guides the policy improvement process. The training details are presented in Appendix \ref{training details}.

\subsection{Neighborhood Optimization}

Before performing the neighborhood optimization process, a sub-ILP is created as defined in (\ref{sub-ilps_format}). Using variable selection probabilities $P(x_{t})$ and the predicted neighborhood size $l_{t}$, we employ a sampling method to select variables for generating the neighborhood $\mathcal{N}_{x_t}$. Based on the ILP instance $\mathcal{Q}$ and $\mathcal{N}_{x_t}$, the sub-ILP $\mathcal{Q}_{t}$ for this step $t$ is formulated and solved within the specified iteration solving time, yielding an updated solution $x_{t+1}$. If the termination conditions are met, the current solution $x_{t+1}$ is returned as the best solution. Otherwise, $x_{t+1}$ is adopted as the current feasible solution, and the solution pool is updated to $X_{t+1} = \{X_{t}, x_{t+1}\}$. The entire process is summarized as an algorithm, shown in Appendix \ref{Algorithm}.
\section{Experiments}
\label{Experiments}

\subsection{Experimental settings}

This section outlines some main experimental settings, including benchmark problems, baseline methods, and performance metrics. More detailed information can be found in Appendix \ref{experiments}.

\textbf{Benchmark problems.} We evaluate HyP-ASO on four widely studied NP-hard ILP benchmarks: Maximum Independent Set (MIS) \cite{tarjan1977finding}, Combinatorial Auction (CA) \cite{devries2003combinatorial}, Minimum Vertex Cover (MVC) \cite{dinur2005hardness}, and Set Covering (SC) \cite{caprara2000setcovering}, following the settings in \cite{Ye23}. For each problem, we construct a small dataset with 100 training instances, 10 validation instances, and 10 test instances. To evaluate generalization to larger problem sizes, we create medium and hard datasets, each consisting of 10 instances. We also assess cross-problem generalization, with results provided in Appendix \ref{Generalization}. The sizes and groupings of all datasets are provided in Appendix \ref{datasets}.


\textbf{Baseline methods.} We compare HyP-ASO with four baseline methods: Gurobi \cite{gurobi}, which is a general solver, and three methods specifically designed for large-scale ILPs based on the LNS framework, including ACP \cite{ye2023adaptive}, GNN\&GBDT \cite{Ye23}, and Light-MILPopt \cite{Ye24}. All methods are evaluated using the same solver settings to ensure a fair comparison and to demonstrate performance improvements over the base solver. Details are provided in Appendix \ref{baselines}. 

\textbf{Performance metrics.} We compare the performance of HyP-ASO with several baseline methods using two criteria: (i) the objective value achieved within the same solving time and (ii) the solving time required to achieve the same objective value. Additionally, we report the average results from five runs for the easy and medium datasets and two runs for the hard dataset to ensure reliability.

\begin{table*}[ht]
\tiny
\setlength{\tabcolsep}{1pt} 
\caption{Comparison of the objective value and solving time on four small datasets: $\text{MIS}_{1}$, $\text{CA}_{1}$, $\text{MVC}_{1}$ and $\text{SC}_{1}$. (a) shows the comparison of the objective value within the same solving time, and (b) shows the comparison of the solving time required to achieve the same objective value. The symbol $>$ indicates that for some instances in the dataset, the maximum solving time was surpassed without reaching the target value. $\uparrow$ and $\downarrow$ denote maximization and minimization problems. Bold values represent the best result. ”-” indicates the experiment was not performed.}
\label{Comparison of objective values and solving time}
\begin{center}
\resizebox{\textwidth}{!}{ 
\begin{tabular}{c@{\hspace{7pt}}c@{\hspace{7pt}}c@{\hspace{7pt}}c@{\hspace{7pt}}c c@{\hspace{7pt}}c@{\hspace{7pt}}c@{\hspace{7pt}}c@{\hspace{7pt}}c}
\multicolumn{5}{c}{\textbf{(a) Comparison of Objective Value}} & \multicolumn{5}{c}{\textbf{(b) Comparison of Solving Time}} \\
\cmidrule(lr){1-5} \cmidrule(lr){6-10} 
Methods & $\text{MIS}_{1}\uparrow$ & $\text{CA}_{1}\uparrow$ & $\text{MVC}_{1}\downarrow$ & $\text{SC}_{1}\downarrow$ & Methods & $\text{MIS}_{1}$ & $\text{CA}_{1}$ & $\text{MVC}_{1}$ & $\text{SC}_{1}$ 
\\ \cmidrule(lr){1-5} \cmidrule(lr){6-10}
Gurobi & 2246.48   & 1399.62   & 2760.49   & 1749.78   & Gurobi & 75.19s & 74.37s & 76.18s & 82.25s \\
ACP & 2303.79$\pm$0.41   & 1401.03$\pm$1.35   & 2682.89$\pm$0.48   & 1592.49$\pm$0.35 & 
ACP & $>$168.99 & $>$253.52 & $>$157.24 & $>$315.94 \\
GNN\&GBDT & 2275.21$\pm$1.00 & \phantom{0} 908.98$\pm$26.40 & 2710.06$\pm$0.50 & 1614.36$\pm$1.69   & GNN\&GBDT & $>$400s & $>$600s & $>$300s & $>$600s \\
Light-MILPopt & - & - & - & - & Light-MILPopt & - & - & - & - \\
\textbf{HyP-ASO} & \textbf{2309.24$\pm$0.97}   & \textbf{1417.71$\pm$0.57}   & \textbf{2677.99$\pm$0.62}   & \textbf{1579.19$\pm$0.77}   & 
\textbf{HyP-ASO} & \textbf{23.77$\pm$1.11s} & \textbf{24.98$\pm$1.17s} & \textbf{20.90$\pm$0.47s} & \textbf{29.34$\pm$1.07s} \\
\cmidrule(lr){1-5} \cmidrule(lr){6-10}  
Time & 40s & 60s & 30s & 60s & Target & 2303.79 & 1401.03 & 2682.89 & 1592.49\\
\cmidrule(lr){1-5} \cmidrule(lr){6-10}
\end{tabular}
}
\end{center}
\vspace{-0.15in} 
\end{table*}

\subsection{Comparison of Objective Value} \label{optimization results}

To ensure a fair comparison, we set the total solving time for each benchmark problem to be the same as in \cite{Ye23} and compare the objective value achieved by different methods within the same solving time. The experimental results are presented in Table \ref{Comparison of objective values and solving time} (a), demonstrating that HyP-ASO significantly outperforms all baseline methods and highlighting the effectiveness of our approach in leveraging an adaptive search optimization framework for large-scale ILPs. This success is primarily attributed to the dynamic adjustments to neighborhood sizes and the efficacy of the customized formula. These mechanisms enable more optimization iterations, promote more effective neighborhood generation, and help avoid local optima, resulting in superior performance compared to baseline methods. Additionally, the Light-MILPopt method was not performed, as it is specially trained on medium datasets and tested on medium and hard datasets, as detailed in Section \ref{generalization results}.

\subsection{Comparison of Solving Time}\label{solving time}

To emphasize the solving efficiency of HyP-ASO, we compare the solving time required to achieve the same objective value across different methods. We set the maximum solving time for each dataset to be 10 times the value in Table \ref{Comparison of objective values and solving time} (a). The objective values achieved by ACP in Table \ref{Comparison of objective values and solving time} (a) serve as the targets. As shown in Table \ref{Comparison of objective values and solving time} (b), HyP-ASO significantly outperforms all baseline methods. It is worth noting that the maximum solving time limits the results. Without this limitation, the solving time of all baselines except Gurobi could exceed current values, which would further expand the gap between baseline methods and ours, highlighting the superior solving efficiency of HyP-ASO.

Moreover, a pseudo-contradiction arises between the results in Table \ref{Comparison of objective values and solving time}. While the objective values from ACP in Table \ref{Comparison of objective values and solving time} (a) serve as the targets, ACP fails to achieve these targets within the same solving time in Table \ref{Comparison of objective values and solving time} (b). This inconsistency is due to the stochastic nature of the ACP method. The results vary across different runs. For example, when solving the small MIS dataset, ACP gets trapped in local optima for certain instances, leading to wasted solving time, which results in the solving time reaching the predefined maximum limit. This explains the appearance of the $>$, and similar situations can also be observed in Section \ref{generalization results}.

\subsection{Generalization Results and Analysis}\label{generalization results}

Methods capable of handling large-scale ILPs are essential, as they are common and offer significant practical value. To evaluate the generalization performance of our approach to larger-scale ILPs, we conduct experiments on medium and hard datasets, with the results presented in Table \ref{Comparison of objective and solving time values medium, hard}. The solving time limits for the medium and hard datasets are set to 300s and 5000s, with the maximum solving time limited to 3000s and 10000s, respectively.

\begin{table*}[htbp]
\scriptsize
\centering
\caption{Comparison of the objective value and solving time on four medium and hard datasets: $\text{MIS}_{2}$, $\text{CA}_{2}$, $\text{MVC}_{2}$, $\text{SC}_{2}$, $\text{MIS}_{3}$, $\text{CA}_{3}$, $\text{MVC}_{3}$, and $\text{SC}_{3}$. The symbol $>$ indicates that for some instances in the dataset, the maximum solving time was surpassed without reaching the target value. $\uparrow$ and $\downarrow$ denote maximization and minimization problems. Bold values represent the best result.}
\label{Comparison of objective and solving time values medium, hard}
\resizebox{\textwidth}{!}{ 
\begin{tabular}{c@{\hspace{7pt}}c@{\hspace{7pt}}c@{\hspace{7pt}}c@{\hspace{7pt}}c@{\hspace{7pt}}c@{\hspace{7pt}}c@{\hspace{7pt}}c@{\hspace{7pt}}c}
\toprule
Methods & $\text{MIS}_{2}\uparrow$ & $\text{CA}_{2}\uparrow$ & $\text{MVC}_{2}\downarrow$ & $\text{SC}_{2}\downarrow$ & $\text{MIS}_{3}\uparrow$ & $\text{CA}_{3}\uparrow$ & $\text{MVC}_{3}\downarrow$ & $\text{SC}_{3}\downarrow$ \\
\midrule
Gurobi & 21800.88 & 12979.83 & 28179.08 & 17939.70 & 218152.86 & 129916.02 & 281928.02 & 320270.24 \\

ACP & 23099.73$\pm$2.19 & 13824.70$\pm$13.42 & 26880.42$\pm$1.64 & 16040.17$\pm$25.34 & 231196.35$\pm$3.10 & 138151.24 & 268868.02$\pm$3.79 & \textbf{161797.14$\pm$5.22} \\

GNN\&GBDT & 22852.36$\pm$5.73 & 11995.92$\pm$16.19 & 27172.18$\pm$6.68 & 16596.90$\pm$34.53 & - & - & - & - \\
Light-MILPopt & \phantom{0}22120.06$\pm$62.92 & - & \phantom{0} 27789.13$\pm$83.10 & 17918.74$\pm$57.15 & 224465.36$\pm$13.64 & - & 275689.44$\pm$244.87 & 172167.26$\pm$63.88 \\

\textbf{HyP-ASO} & \textbf{23135.74$\pm$4.62} & \textbf{14052.91$\pm$9.68} & \textbf{26857.15$\pm$9.53} & \textbf{16012.09$\pm$22.21} & \textbf{231633.02$\pm$97.54} & \textbf{139998.55$\pm$329.66} & \textbf{268563.22$\pm$47.42} & 162807.21$\pm$107.07 \\

\midrule
Time & 300s & 300s & 300s & 300s & 5000s & 5000s & 5000s & 5000s \\
\bottomrule
\toprule
Gurobi & $>$3000s & $>$3000s & $>$3000s & $>$3000s & $>$10000s & $>$10000s & $>$10000s & $>$10000s \\
ACP &  $>$2150.18s & $>$2143.25s & $>$1286.89s & $>$1327.72s & $>$7958.60s & $>$6045.12s & $>$7308.04s & $>$5801.19s \\
GNN\&GBDT & $>$3000s & $>$3000s & $>$3000s & $>$3000s & - & - & - & - \\
Light-MILPopt & $>$3000s & - & $>$3000s & $>$3000s & >10000s & - & >10000s & >10000s \\
\textbf{HyP-ASO} & \textbf{218.86$\pm$15.04}s & \textbf{146.61$\pm$8.69}s & \textbf{254.25$\pm$20.26}s & \textbf{257.41$\pm$16.57}s & \textbf{2940.49$\pm$46.73}s & \textbf{2269.55$\pm$156.32}s & \textbf{3473.25$\pm$140.08}s & \textbf{6517.22$\pm$103.34}s \\
\midrule
Target & 23099.73 & 13824.70 & 26880.42 & 16040.17 & 231196.35 & 138151.24 & 268868.02 & 161797.14 \\
\bottomrule
\end{tabular}
}
\end{table*}

As shown in Table \ref{Comparison of objective and solving time values medium, hard}, HyP-ASO significantly outperforms all baseline methods across four medium datasets. Compared to Gurobi, our approach reduces the solving time by over 90\%. For hard datasets, HyP-ASO outperforms all baseline methods except for the SC dataset. However, HyP-ASO can achieve the targets within the maximum solving time, demonstrating its outstanding ability to escape local optima. In addition, we could not complete tests for the GNN\&GBDT method on the hard datasets due to server GPU memory limitations, as it requires more than 24GB of RAM. The above results and findings suggest that our approach is lightweight and highly scalable.

\begin{table*}
\vspace{-1em}  
\centering
\tiny
\caption{Comparison of different neighborhood size strategies in LNS methods.}
\begin{tabular}{c@{\hspace{10pt}}c@{\hspace{10pt}}c@{\hspace{10pt}}c@{\hspace{10pt}}c}
\toprule
Methods & $\text{MIS}_{1}\uparrow$ & $\text{CA}_{1}\uparrow$ & $\text{MVC}_{1}\downarrow$ & $\text{SC}_{1}\downarrow$ \\
\midrule
HyP-ASO-G & 2233.50$\pm$6.99 & 1368.22$\pm$9.08 & 2748.57$\pm$6.27 & 1646.02$\pm$8.74 \\
HyP-ASO-B & 2252.21$\pm$3.56 & 1402.37$\pm$1.15 & 2731.50$\pm$4.03 & 1619.32$\pm$2.20 \\
HyP-ASO-U & 2291.77$\pm$3.77 & 1408.07$\pm$2.75 & 2698.73$\pm$4.36 & 1589.48$\pm$5.54 \\
HyP-ASO-F30\% & 2221.74$\pm$1.80 & 1363.17$\pm$2.08 & 2770.27$\pm$1.19 & 1675.64$\pm$1.96 \\
HyP-ASO-F50\% & 2274.32$\pm$1.50 & 1406.47$\pm$0.58 & 2719.10$\pm$1.55 & 1603.90$\pm$1.45 \\
HyP-ASO-F70\% & 2305.07$\pm$0.51 & 1415.05$\pm$0.28 & 2682.01$\pm$1.00 & \textbf{1578.12$\pm$0.37} \\
\textbf{HyP-ASO} & \textbf{2309.24$\pm$0.97}   & \textbf{1417.71$\pm$0.57}   & \textbf{2677.99$\pm$0.62}   & 1579.19$\pm$0.77   \\
\midrule
Time & 40s & 60s & 30s & 60s \\
\bottomrule
\label{Ablation experiments}
\end{tabular}
\vspace{-3em} 
\end{table*}

\subsection{Ablation Experiments}





To evaluate the effectiveness of the RL policy, we conduct ablation experiments from two perspectives: (1) replacing the learned RL policy for neighborhood size prediction with common statistical distributions, Gaussian, Uniform, and Beta, and (2) adopting fixed-size neighborhood strategies with the size set to 30\%, 50\%, or 70\% of the total variables. These variants are denoted as HyP-ASO-G, HyP-ASO-U, HyP-ASO-B, and HyP-ASO-F30\%, HyP-ASO-F50\%, HyP-ASO-F70\%, respectively. In all cases, the proposed variable selection mechanism is retained.

\begin{wraptable}{r}{0.45\linewidth}
\vspace{-1em}  
    \centering
    \tiny
    \caption{Comparison of different variable selection strategies in LNS methods for the SC problem.}
    \begin{tabular}
        {c@{\hspace{1pt}}c@{\hspace{1pt}}c@{\hspace{1pt}}c}
        \toprule
        Methods & $\text{SC}_{1}\downarrow$ & $\text{SC}_{2}\downarrow$ & $\text{SC}_{3}\downarrow$ \\
        \midrule
        HyP-ASO-UF & 1580.68$\pm$0.54 & 16041.14$\pm$17.31& 163486.81$\pm$100.28 \\
        HyP-ASO & \textbf{1579.19$\pm$0.77} & \textbf{16012.09$\pm$22.21} & \textbf{162807.21$\pm$107.07}  \\
        \midrule
        Time & 60s & 300s & 5000s \\
        \bottomrule
    \label{fig: abl2}
    \end{tabular}
    \vspace{-2em}  
\end{wraptable}

Additionally, we evaluate the impact of the variable selection mechanism by comparing it with a uniform selection strategy, where all variables have equal probability. This experiment uses the same sampling method but replaces the customized formula with the uniform weight $\frac{1}{n}$, called HyP-ASO-UF.

The results in Table \ref{Ablation experiments} demonstrate that the RL-based policy consistently performs best among all methods, highlighting the critical role of learning adaptive neighborhood sizes. Moreover, Table \ref{fig: abl2} shows that replacing the customized formula with the uniform selection strategy leads to a performance degradation, indicating that the design of the variable selection mechanism is also beneficial. Furthermore, this approach offers a new and valuable perspective for generating neighborhoods in LNS for large-scale ILPs.

\subsection{Convergence analysis}

\begin{wrapfigure}{r}{0.45\linewidth}
\vspace{-2em}  
\centering
\includegraphics[width=1\linewidth]{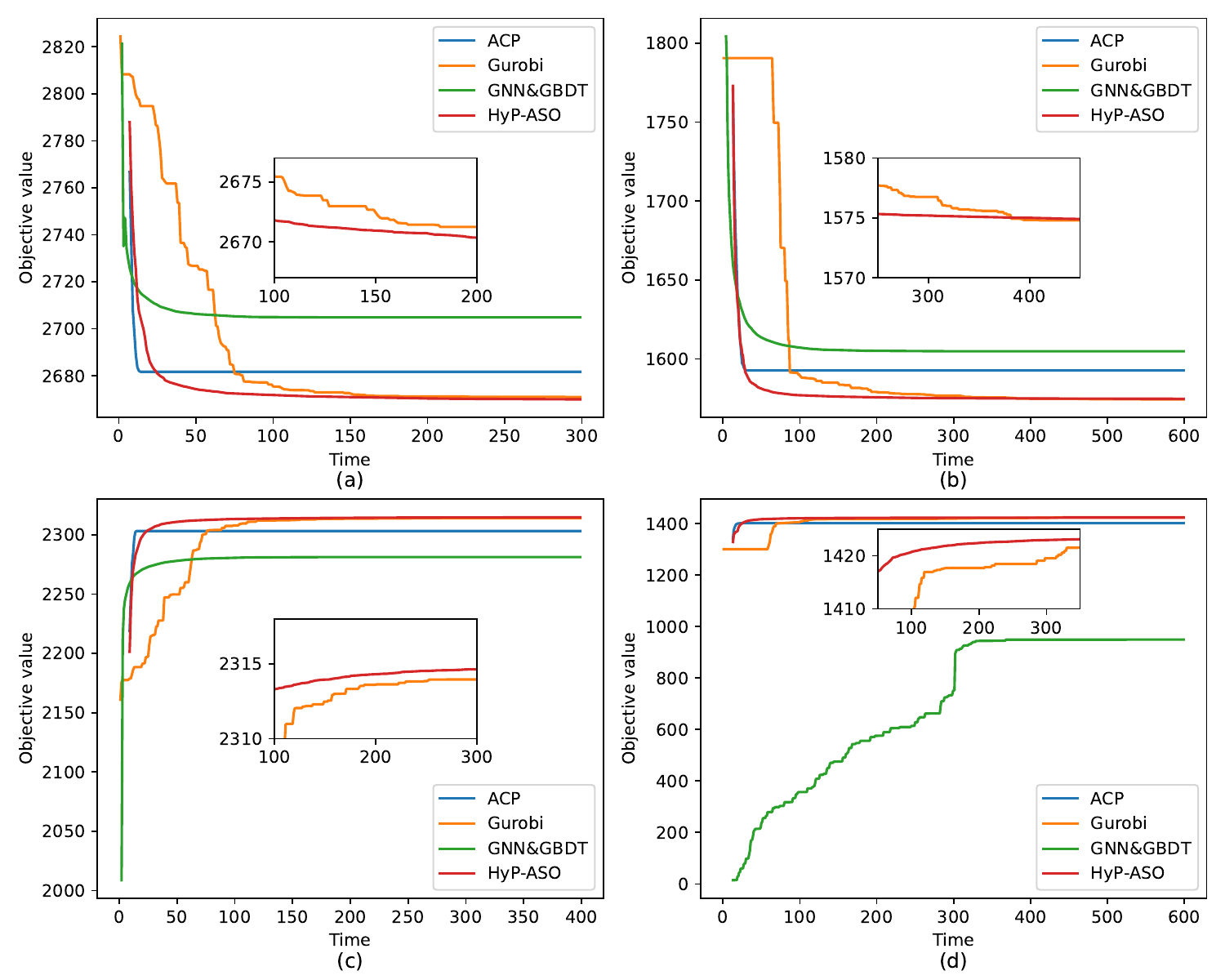}
\caption{Objective values over time for four benchmark ILPs: (a) MVC, (b) SC, (c) IS, and (d) CA.}
\label{fig: easy_plot}
\vspace{-3em}  
\end{wrapfigure}

Convergence is a critical metric for evaluating the search optimization framework. To assess this, we set the total solving time for the four datasets to 10 times the values in Table \ref{Comparison of objective values and solving time} (a) and recorded the changes in the objective value over time achieved by each method. The results are presented in
Figure \ref{fig: easy_plot}, showing that our approach converges more rapidly to higher-quality objective values than others, with its final performance comparable to Gurobi. The convergence results on medium and hard datasets are provided in Appendix \ref{Convergence analysis}.


\section{Conclusion, Limitation, and Future Work}
\label{Conclusion}

This paper proposes HyP-ASO, a hybrid policy-based adaptive search optimization framework. It employs a customized formula based on a solution pool to compute variable selection probabilities and uses an RL policy to predict neighborhood sizes. Variables are then sampled to form neighborhoods and generate sub-ILPs, whose solutions update both the current solution and the pool. Experimental results show that HyP-ASO outperforms all baseline methods on four benchmark ILPs. 

\textbf{Limitations.} In all experiments, the solving time is treated as a fixed and manually tuned hyperparameter, consistent with prior works. Moreover, the initial feasible solution is obtained quickly by the solver. These settings indicate that the current approach is not fully automated. In addition, the customized scoring formula heavily depends on the availability of high-quality feasible solutions, which may limit its robustness in more challenging scenarios.

\textbf{Future Work.} In future work, we aim to develop an automated mechanism for allocating solving time and dynamically managing the feasible solution pool to improve adaptability and performance. We also plan to develop a more efficient variable selection mechanism by machine learning to better handle complex scenarios, increasing its generality and reliability. Furthermore, we will investigate the applicability of HyP-ASO to general ILP problems, particularly those involving complex constraint structures.

\newpage
\newpage
\bibliography{reference}
\newpage
\newpage
\appendix

\section{Algorithm}\label{Algorithm}
This paper has provided an overview of the proposed framework and detailed its key components in Section \ref{method}. To further clarify the complete process of HyP-ASO, we present an algorithm that summarizes the procedure. Given an instance $\mathcal{Q}$, we initialize the solution ${x}_{0}$, the solution pool $X_{0}$, the number of decision variables $n$, the bipartite graph $\mathcal{G} = (\mathcal{V}, \mathcal{C}, \mathcal{E})$, the initial solving time $t_{0}$, the iteration solving time $t_{p}$, the total solving time $t_{total}$, the maximum number of steps $T$, the customized formula (\ref{customized_formula}), and the policy $\pi_{\theta}(a_{t}|s_{t})$. The goal is to find a high-quality feasible solution within the specified total solving time $t_{total}$. The algorithmic process is outlined below.
\begin{algorithm}[ht]
   \caption{HyP-ASO Algorithm}
   \label{alg: HyP-ASO}
\begin{algorithmic}
    \STATE {\bfseries Input:} ILP instance $\mathcal{Q}$, number of decision variables $n$, bipartite graph $\mathcal{G} = (\mathcal{V}, \mathcal{C}, \mathcal{E})$, initial solving time $t_{0}$, iteration solving time $t_{p}$, total solving time $t_{total}$, maximum steps $T$, customized formula (\ref{customized_formula}), policy $\pi_{\theta}(a_{t}|s_{t})$
    \STATE {\bfseries Output:} The best solution ${x}_{t}$ found within the total solving time $t_{total}$
    \STATE {\bfseries Initialization:}
    \STATE \quad Record the start time $t_{start}$
    \STATE \quad Initialize feasible solution ${x}_{0}$ and the solution pool $X_{0}$ 
    \STATE \quad Initialize $\mathcal{G}$ features $(\mathbf{V}, \mathbf{C}, \mathbf{E})$
    \STATE \quad Set the step counter $t = 0$
    \WHILE{$t \leq T$}
        \STATE Calculate variable selection probabilities $P(x_{t})$ using the formula (\ref{customized_formula}) and current solution pool $X_{t}$
        \STATE Predict neighborhood size $l_{t} = a_{t} \times n$ using the policy $\pi_{\theta}$
        \STATE Sample $l_{t}$ variables to generate the neighborhood $\mathcal{N}({x_t})$
        \STATE Fix variables $x_{i} \notin \mathcal{N}({x_t})$, denoted as $F_{var}$
        \STATE Unfix variables $x_{i} \in \mathcal{N}({x_t})$, denoted as $U_{var}$
        \STATE Solve the sub-ILP $\mathcal{Q}_{t}(\mathcal{Q}, F_{var}, U_{var})$ to obtain ${x}_{t+1}$
        \STATE Record the end time $t_{end}$
        \IF{$t_{end} - t_{start} \leq t_{total}$}
            \STATE Update the current solution ${x}_{t+1}$ as the best solution
            \STATE Update the solution pool $X_{t+1} = \{X_{t}, \textbf{x}_{t+1}\}$
            \STATE Update variable features $\mathbf{V} = \{\mathbf{V}, \textbf{x}_{t+1}\}$
            \STATE Increment the step counter $t = t + 1$
        \ELSE
            \STATE Exit the loop
        \ENDIF
    \ENDWHILE
    \STATE {\bfseries Return:} The best solution obtained so far, ${x}_{t+1}$
\end{algorithmic}
\end{algorithm}

\section{Experimental Details} \label{experiments} 
\subsection{Experimental settings} \label{experiment details} 
All experiments were run on a machine equipped with an Intel(R) Xeon(R) Gold 5220 CPU @ 2.20 GHz and NVIDIA TITAN RTX GPUs, each with 24 GB of RAM.
Gurobi 11.0.3 \cite{gurobi} is used as the solver, while our model is implemented with PyTorch 2.2.0 \cite{pytorch2024} and PyG 2.5.2 \cite{pyg2024}.

While we use Gurobi \cite{gurobi} as the primary solver, SCIP \cite{Bolusani2024} and other solvers are also available. To verify this, we conduct experiments using SCIP as the primary solver in Appendix \ref{SCIP_experiment}. For each benchmark problem, we employ the Adam optimizer with an initial learning rate of $3 \times 10^{-4}$, applying a linear decay strategy, with a minimum learning rate set to $1 \times 10^{-5}$. We set the discount factor $k_r$ of the cumulative reward (\ref{return}) to $0.01$. 

In addition, the initial solving time and iteration solving time for solving instances in each dataset are shown in Table \ref{solving time record}. For experiments that compare the solving time, the allocation method of the solving time settings remains unchanged, with only the total solving time being extended to 10 times the original solving time.

\begin{table}[ht]
\caption{Allocation of the solving time. S.T. represents the solving time.}
\label{solving time record}
\vskip 0.1in 
\begin{small}
\begin{center}
\begin{tabular}{c@{\hspace{10pt}}c@{\hspace{10pt}}c@{\hspace{10pt}}c@{\hspace{10pt}}c}
\toprule
Datasets & Size & Initial S.T. & Iteration S.T. & Total S.T. \\
\midrule
 & small & 8s & 4s & 40s \\
\textbf{MIS} & medium & 60s & 30s & 300s\\
 & hard & 600s & 300s & 5000s\\
\midrule
 & small & 12s & 6s & 60s \\
\textbf{CA} & medium & 60s & 30s & 300s\\
 & hard & 600s & 300s & 5000s\\
\midrule
 & small & 6s & 3s & 30s \\
\textbf{MVC} & medium & 60s & 30s & 300s\\
 & hard & 600s & 300s & 5000s\\
\midrule
 & small & 12s & 6s & 60s \\
\textbf{SC} & medium & 60s & 30s & 300s\\
 & hard & 600s & 300s & 5000s\\
\bottomrule
\end{tabular}
\end{center}
\end{small}
\vspace{-0.1 in} 
\end{table}

\subsection{Datasets}\label{datasets}

For four widely used NP-hard benchmark ILPs, the existing datasets cannot meet such large-scale data requirements, so we use data generators to generate training and test datasets. Specifically, for the Maximum Independent Set problem (MIS) or Minimum Vertex Covering problem (MVC) with $n$ decision variables and $m$ constraints, we generate a random graph with $n$ nodes and $m$ edges to correspond to an ILP instance that meets the scale requirements. For the Combinatorial Auction problem (CA) with $n$ decision variables and $m$ constraints, we generate a random problem with $n$ items and $m$ bids, where each bid includes 5 items. For the Set Covering problem (SC) with $n$ decision variables and $m$ constraints, we generate a random problem with $n$ items and $m$ sets where each set bid includes 4 items. The specific sizes and groupings of instances for all ILP datasets are provided in Table \ref{instance size table}.

\begin{table}[ht]
\caption{Instance sizes for benchmark ILPs. Num. of Vars and Num. of Cons mean the number of variables and constraints, respectively.}
\label{instance size table}
\vskip 0.1in 
\begin{small}
\begin{center}
\begin{tabular}{c@{\hspace{10pt}}c@{\hspace{10pt}}c@{\hspace{10pt}}c@{\hspace{10pt}}c}
\toprule
Datasets & Size & Num. of Vars & Num. of Cons \\
\midrule
 & small & 10,000 & 30,000 \\
\textbf{MIS} & medium & 100,000 & 300,000 \\
 & hard & 1,000,000 & 3,000,000 \\
\midrule
 & small & 10,000 & 10,000 \\
\textbf{CA} & medium & 100,000 & 100,000 \\
 & hard & 1,000,000 & 1,000,000 \\
\midrule
 & small & 10,000 & 30,000 \\
\textbf{MVC} & medium & 100,000 & 300,000 \\
 & hard & 1,000,000 & 3,000,000 \\
\midrule
 & small & 20,000 & 20,000 \\
\textbf{SC} & medium & 200,000 & 200,000 \\
 & hard & 2,000,000 & 2,000,000 \\
\bottomrule
\end{tabular}
\end{center}
\end{small}
\vspace{-0.1 in} 
\end{table}

\subsection{Baselines}\label{baselines}
\textbf{ACP:} The method considers the correlation between variables in ILPs. It starts with an initial feasible solution. Then, it randomly partitions the constraints into blocks, where the number of blocks is adaptively adjusted to avoid local optima. Finally, it uses a subroutine solver to optimize the decision variables in a randomly selected block of constraints to obtain a better feasible solution. 

However, although ACP can achieve promising performance in a shorter time, it is prone to getting trapped in local optima. This is due to the decrease in the number of blocks, which causes an increase in the number of variables, making iteration optimization more difficult and limiting the upper bound of its performance.

In our experiments, we observe this issue from three perspectives. First, the performance of HyP-ASO outperforms ACP, suggesting that the optimization capacity of HyP-ASO has a higher upper bound. Second, ACP achieves a better average objective value than HyP-ASO on the hard SC dataset within 5000s. However, when using the objective values obtained by ACP as the target and resolving the instances, we find that ACP fails to reach the target value within 10,000s. In contrast, for some instances on the SC dataset, our method consistently achieves the target value within the same time limit. This discrepancy is due to the inherent stochastic nature of ACP's solving process, which means that the second solving performance does not always surpass the first for each instance in the dataset. In addition, if ACP becomes trapped in a local optimum, it struggles to escape, as solving later-stage subproblems becomes increasingly time-consuming. In contrast, our method demonstrates a stronger ability to escape local optima. Third, the convergence analysis further supports the robustness of HyP-ASO, showing that HyP-ASO is more stable and consistently outperforms ACP across different benchmark problems. 

\textbf{GNN\&GBDT and Light-MILPopt:} These two methods leverage ML to predict initial feasible solutions and generate subproblems through neighborhood partition, which are solved using a small-scale optimizer. GNN\&GBDT can be divided into three stages: Multitask GNN Embedding, GBDT Prediction, and Neighborhood Optimization. In the stage of Multi-task GNN Embedding, the ILP is represented as a bipartite graph, followed by a graph partition algorithm to divide the bipartite graph into blocks. Then, a multi-task GNN with half convolutions is used to learn the embedding of decision variables, where the loss function is a metric for both the optimal solution and the graph partition. In the stage GBDT prediction, the GBDT is used to predict the optimal value of the decision variable in the IP through the variable embedding. In the stage of Neighborhood Optimization, some decision variables are fixed as the rounding results of the predicted values of GBDT, and a search with a fixed radius is used to find an initial solution. Then, neighborhood search and neighborhood crossover are used iteratively to improve the current solution under the guidance of the neighborhood partition.  

Light-MILPopt can be divided into four stages: Firstly, the ILP is represented as a bipartite graph, and the FENNEL graph partitioning algorithm is employed for problem division to reduce model computational costs. Secondly, given the divided graph representations, the Edge Aggregated Graph Attention Network (EGAT), trained by a small-scale training dataset, is used for predicting and generating the initial solution of the original large-scale ILP. Thirdly, the predicted solution introduces a decision variable confidence threshold and a KNN-based strategy for constraints, which aims at problem reduction. Finally, based on problem division and reduction, the subgraph clustering and active constraint updating guide neighborhood search and individual crossover, iteratively improving the current solution by a lightweight optimizer.

However, these two methods are also prone to getting trapped in local optima. While they perform well in obtaining the initial solution, they lack flexibility and effectiveness in neighborhood generation. This limitation arises from their reliance on a fixed neighborhood size and imprecise rules for choosing neighborhood variables, preventing them from fully leveraging the potential of LNS. Experimental results show that, when solving large-scale ILPs, these methods perform even worse than the ACP in a short time.

\subsection{Training Details}\label{training details}

This paper uses PPO to train the policy network, with parameters listed in Table \ref{PPO parameters}. The learning rate follows a linear decay strategy: $\text{lr}(t) = \max\big(1-\frac{t}{T}\times \text{lr}_0, \text{lr}_{\min}\big),$ where $t$ represents the current iteration, $\text{lr}_0$ is the initial learning rate, $\text{lr}_{\min}$ is the minimum learning rate, and $T$ is the total training iterations. The PPO implementation follows \cite{huang2022cleanrl}.
\begin{table}[ht]
\caption{The main parameters of the PPO algorithm used in our implementation.}
\label{PPO parameters}
\vskip 0.1in 
\begin{small}
\begin{center}
\begin{tabular}{l@{\hspace{10pt}}|c@{\hspace{10pt}}}
\toprule
Parameters & Value \\
\midrule
Optimizer & Adam \\
Learning Rate & 3e-4 \\
Number of environment & 5 \\
Number of epochs & 5 \\
Number of training iterations & 500 \\
Batchsize & 80 \\
Minibatchsize & 20 \\
Number of iteration step & 16 \\
Update epochs & 5 \\
GAE parameter & 0.95 \\
Clip coefficient & 0.2 \\
VF coefficient & 0.5 \\
ENT coefficient & 0.02 \\
\bottomrule
\end{tabular}
\end{center}
\end{small}
\vspace{-0.1 in} 
\end{table}

\subsection{Network Structure}\label{network structure}

\begin{figure}[ht]
    \centering
    \includegraphics[width=0.95\linewidth]{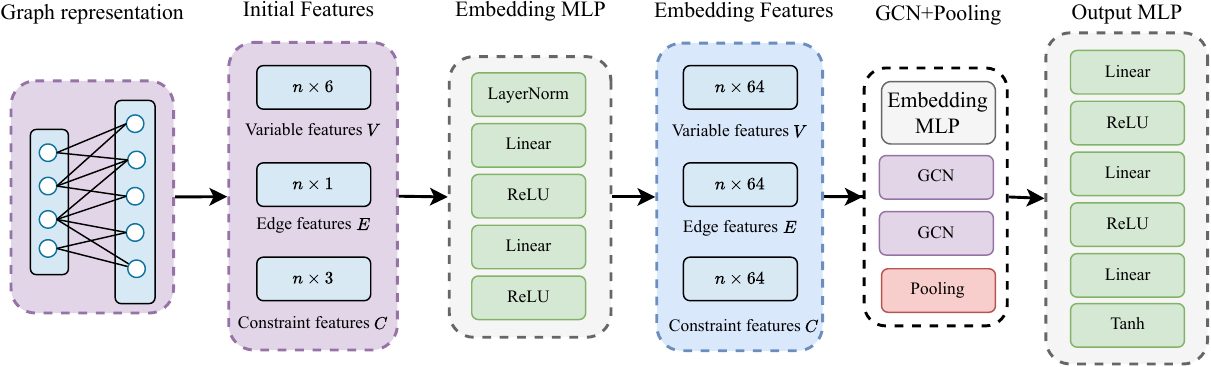}
    \caption{The network structure of the policy in our framework.}
    \label{fig: network structure}
\end{figure}

In this paper, we propose using PPO to train the policy, with the policy network structure illustrated in Figure \ref{fig: network structure}. Given the graph representation of an instance, the network outputs the mean and standard deviation, which define the policy distribution. Actions are then sampled from this distribution. For simplicity, the figure only shows a single output MLP layer. However, it consists of two separate MLP layers: one for the mean and another for the standard deviation. 

It is worth noting that the global pooling layer is designed to facilitate generalization across ILPs of varying sizes. ReLU introduces non-linearity to stabilize the representation learning, while Tanh ensures that the mean and standard deviation outputs remain constrained within the range $(0, 1)$.

\section{Generalization Performance across Different Problems}\label{Generalization}

We hope to establish a general framework that can be applied across various problems. To evaluate the generalization performance of HyP-ASO across different problems, we conduct experiments where the model is trained on one dataset and tested on others. The experimental results are presented in Table \ref{Generalization Performance across Different Problems}. Remarkably, the model trained on the MIS dataset demonstrates the best generalization ability, outperforming all baseline methods across four benchmark ILPs. This suggests the model has strong robustness. It can be effectively transferred to other problems. Furthermore, the consistency of its performance across different problems underscores the flexibility and generalizability of HyP-ASO.

\begin{table}[ht]
\caption{The generalization results across different problems. Comparison of the objective value within the same solving time on four small datasets. $\uparrow$ and $\downarrow$ denote maximization and minimization problems. Bold values represent the best result. For each value, the dataset corresponding to the row is the training dataset, and the dataset corresponding to the column is the test dataset. For example, 1413.76 means the model is trained on the MIS dataset and tested on the CA dataset.}
\label{Generalization Performance across Different Problems}
\begin{center}
\begin{tabular}{c@{\hspace{10pt}}c@{\hspace{10pt}}c@{\hspace{10pt}}c@{\hspace{10pt}}c}
\toprule
 & $\text{MIS}_{1}\uparrow$ & $\text{CA}_{1}\uparrow$ & $\text{MVC}_{1}\downarrow$ & $\text{SC}_{1}\downarrow$ \\
\midrule
$\text{MIS}_{1}$ & \textbf{2309.24} & 1413.76 & 2682.26 & 1582.83 \\
$\text{CA}_{1}$ & 2301.66 & \textbf{1417.71} & 2687.97 & 1581.48 \\
$\text{MVC}_{1}$ & 2306.26 & 1410.21 & \textbf{2677.99} & 1584.33 \\
$\text{SC}_{1}$ & 2301.62 & 1413.46 & 2687.86 & \textbf{1579.19}   \\
\midrule
Time & 40s & 60s & 30s & 60s \\
\bottomrule
\end{tabular}
\end{center}
\vspace{-0.1in} 
\end{table}

\section{Generalization for MIPLIB 2017}

To further evaluate the generalization performance of HyP-ASO and its potential for real-world applications, we test the model trained on the SC dataset using instances from the MIPLIB benchmark \cite{Gleixner2021Miplib}. Specifically, we select the \textit{scpk4} and \textit{scpn2} instances. The \textit{scpk4} instance contains 100,000 variables and 2,000 constraints, while the \textit{scpn2} instance comprises 1,000,000 variables and 5,000 constraints. These two instances correspond to medium and hard problem settings, respectively. It is important to note that \textit{scpk4} and \textit{scpn2} are SC problems. 
However, their distributions differ from the SC dataset used in this paper.

We compare the objective value achieved by HyP-ASO with those obtained using Gurobi and Light-MILPopt within the same solving time. The solving time for \textit{scpk4} is set to match that of the medium dataset used in this paper, while the solving time for \textit{scpn2} follows the setting outlined in \cite{Ye24}. The results presented in Table \ref{MIPLIB2017} highlight the strong generalization capability of HyP-ASO.

\begin{table}[ht]
\caption{The generalization results for instances \textit{scpk4} and \textit{scpn2} in MIPLIB 2017. Comparison of the objective value within the same solving time on instances \textit{scpk4} and \textit{scpn2}. $\downarrow$ denotes minimization problems. Bold values represent the best result.}
\label{MIPLIB2017}
\begin{center}
\begin{tabular}{c@{\hspace{10pt}}c@{\hspace{10pt}}c@{\hspace{10pt}}c@{\hspace{10pt}}c}
\toprule
Methods & $\text{scpk4}\downarrow$ & $\text{scpn2}\downarrow$\\
\midrule
Gurobi & 328 & 700 \\ 
Light-MILPopt & 452 & 7895 \\
\textbf{HyP-ASO} & \textbf{324} & \textbf{661}\\ 
\midrule
Time & 300s & 2000s\\
\bottomrule
\end{tabular}
\end{center}
\vspace{-0.1in} 
\end{table}

\section{SCIP as the Primary Solver}\label{SCIP_experiment}

In this section, SCIP is used as the primary solver to conduct experiments on four benchmark ILPs. We compare the objective value obtained by HyP-ASO, SCIP, and ACP within the same solving time. The experimental settings are consistent with those outlined in Section \ref{optimization results}, and the results are presented in Table \ref{SCIP_results}. These results demonstrate that HyP-ASO outperforms other methods and confirm the versatility of our approach in utilizing different solvers as the primary solvers.

\begin{table}[ht]
\caption{The comparison results of using SCIP as the primary solver. Comparison of the objective value within the same solving time on four benchmark ILPs. $\uparrow$ and $\downarrow$ denote maximization and minimization problems. Bold values represent the best result.}
\label{SCIP_results}
\vskip -0.1in 
\begin{center}
\begin{small}
\begin{tabular}{c@{\hspace{10pt}}c@{\hspace{10pt}}c@{\hspace{10pt}}c@{\hspace{10pt}}c}
\toprule
Methods & $\text{MIS}_{1}\uparrow$ & $\text{CA}_{1}\uparrow$ & $\text{MVC}_{1}\downarrow$ & $\text{SC}_{1}\downarrow$ \\
\midrule
SCIP & 1867.70 & 1120.98 & 3117.61 & 2512.82 \\
ACP & 2228.57 & 1374.68 & 2758.43 & 1640.50 \\
\textbf{HyP-ASO} & \textbf{2289.93}   & \textbf{1388.61}   & \textbf{2706.35}   & \textbf{1619.67}   \\
\midrule
Time & 40s & 60s & 30s & 60s \\
\bottomrule
\end{tabular}
\end{small}
\end{center}
\vspace{-0.1in} 
\end{table}

\section{Convergence Analysis}\label{Convergence analysis}

In this section, we extend the convergence analysis to the medium and hard datasets to further evaluate the performance of our approach. The experiments are conducted using the same settings as Section \ref{solving time}. Specifically, the total solving time for the medium datasets is increased to 10 times the values in Table \ref{Comparison of objective and solving time values medium, hard}, and for the hard datasets, it is extended by 2 times. We record the changes in the objective value over time to assess the convergence behavior of each method. 

\begin{figure*}[ht]
    \centering
    \begin{minipage}[t]{0.48\linewidth}
        \centering
        \includegraphics[width=\linewidth]{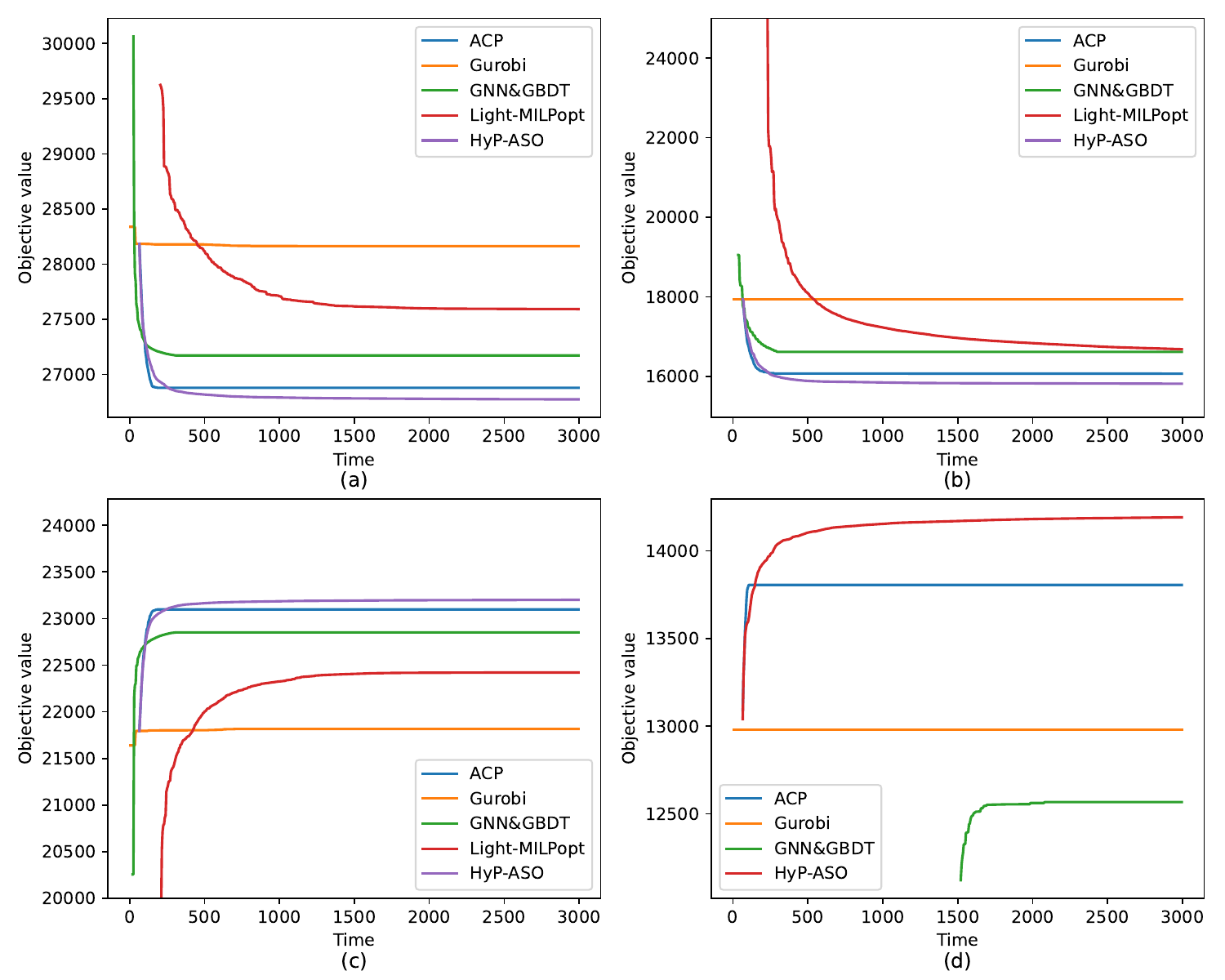}
        \caption{Objective values over time for medium benchmark ILPs: (a) MVC, (b) SC, (c) IS, and (d) CA.}
        \label{fig: medium_plot}
    \end{minipage}
    \hfill
    \begin{minipage}[t]{0.48\linewidth}
        \centering
        \includegraphics[width=\linewidth]{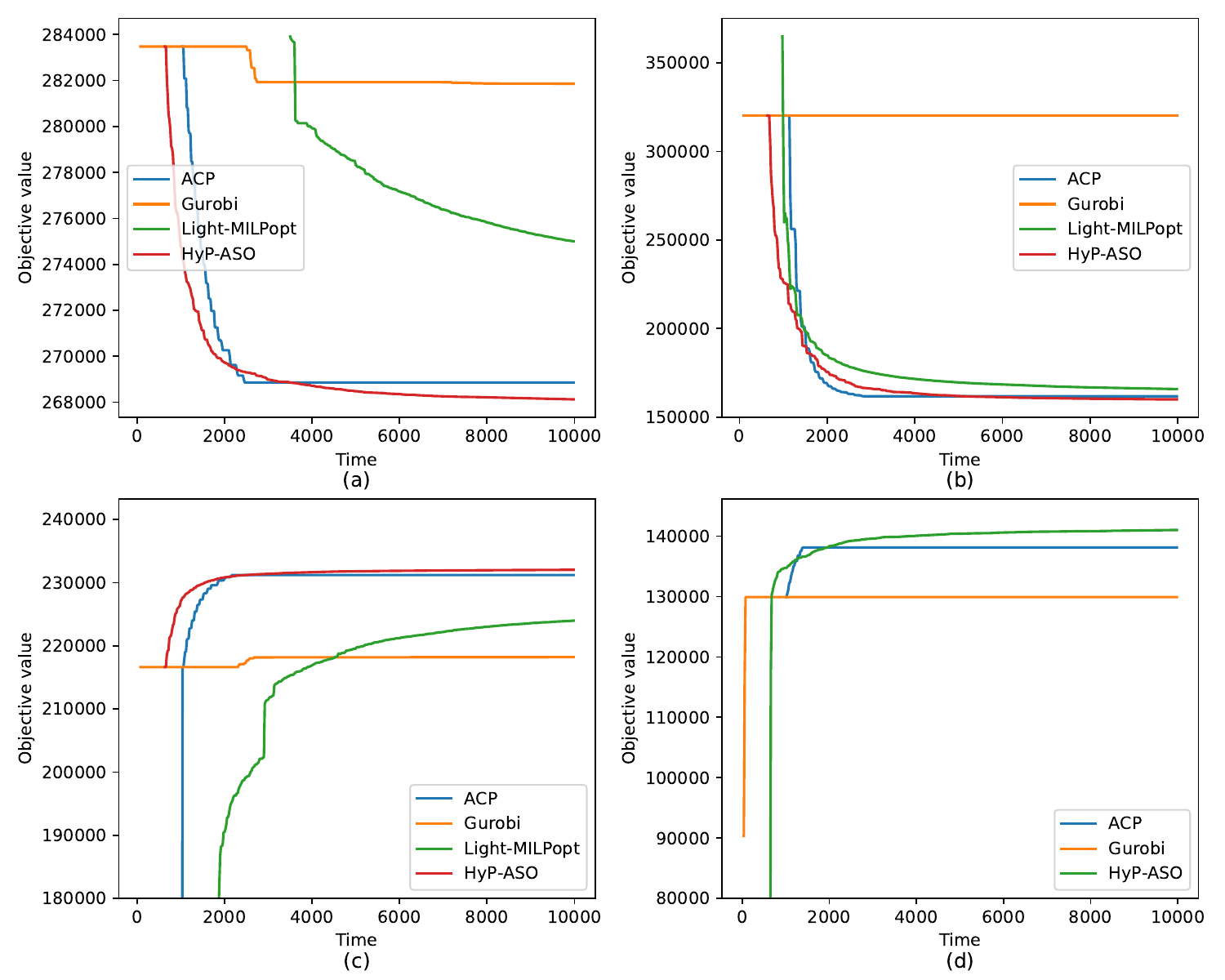}
        \caption{Objective values over time for hard benchmark ILPs: (a) MVC, (b) SC, (c) IS, and (d) CA.}
        \label{fig: hard_plot}
    \end{minipage}
\end{figure*}

The results are presented in Figures \ref{fig: medium_plot} and \ref{fig: hard_plot}. We can observe that HyP-ASO consistently converges to the best objective value compared to all baseline methods across four benchmark ILPs. These results further validate the good performance and superior convergence properties of HyP-ASO. With its ability to rapidly obtain high-quality solutions and maintain stability, HyP-ASO is proven to be exceptionally effective for solving large-scale ILPs.

\newpage
\end{document}